\documentclass{article}

\pdfoutput=1

\usepackage{microtype}
\usepackage{graphicx}
\usepackage{subfigure}
\usepackage{booktabs} % for professional tables

\usepackage{hyperref}

\usepackage[accepted]{icml2024}

\usepackage{amsmath}
\usepackage{amssymb}
\usepackage{mathtools}
\usepackage{amsthm}

\usepackage[capitalize,noabbrev]{cleveref}

\usepackage{amsfonts}
\usepackage{textcomp}
\usepackage{xcolor}
\usepackage{hhline}
\usepackage{colortbl}
\usepackage{multirow}
\usepackage{algorithm}
\usepackage{algorithmic}
\hypersetup{
    colorlinks=true,
    linkcolor=red,
    filecolor=green,      
    urlcolor=blue,
    pdftitle={DeBUGCN - Vartak et. al.},
    }

\theoremstyle{plain}

\theoremstyle{definition}

\theoremstyle{remark}

\usepackage[textsize=tiny]{todonotes}

\icmltitlerunning{DeBUGCN - Detecting Backdoors in CNNs Using Graph Convolutional Networks}

\begin{document}

\twocolumn[
\icmltitle{DeBUGCN - Detecting Backdoors in CNNs Using\\Graph Convolutional Networks}



\icmlsetsymbol{equal}{*}

\begin{icmlauthorlist}
\icmlauthor{Akash Vartak}{xxx}
\icmlauthor{Khondoker Murad Hossain}{yyy}
\icmlauthor{Tim Oates}{zzz}

\icmlauthoremail{akashvartak@umbc.edu}{xxx}
\icmlauthoremail{hossain10@umbc.edu}{yyy}
\icmlauthoremail{oates@cs.umbc.edu}{zzz}
\end{icmlauthorlist}

\icmlaffiliation{xxx}{}
\icmlaffiliation{yyy}{}
\icmlaffiliation{zzz}{Department of Computer Science \& Engineering, University of Maryland Baltimore County, Baltimore, USA}

\icmlcorrespondingauthor{Akash Vartak}{akashvartak@umbc.edu}

\icmlkeywords{Machine Learning, ICML}

\vskip 0.3in
]



\printAffiliationsAndNotice{}  % leave blank if no need to mention equal contribution

\begin{abstract}
    Deep neural networks (DNNs) are becoming commonplace in critical applications, making their susceptibility to backdoor (trojan) attacks a significant problem. In this paper, we introduce a novel backdoor attack detection pipeline, detecting attacked models using graph convolution networks (DeBUGCN). 
    To the best of our knowledge, ours is the first use of GCNs for trojan detection. We use the static weights of a DNN to create a graph structure of its layers. A GCN is then used as a binary classifier on these graphs, yielding a trojan or clean determination for the DNN. 
    To demonstrate the efficacy of our pipeline, we train hundreds of clean and trojaned CNN models on the MNIST handwritten digits and CIFAR-10 image datasets, and show the DNN classification results using DeBUGCN. For a true In-the-Wild use case, our pipeline is evaluated on the TrojAI dataset which consists of various CNN architectures, thus showing the robustness and model-agnostic behaviour of DeBUGCN. 
    Furthermore, on comparing our results on several datasets with state-of-the-art trojan detection algorithms, DeBUGCN is faster and more accurate.
\end{abstract}

\section{Introduction}
\label{sec:introduction}

    Deep neural networks (DNNs) have achieved state-of-the-art performance on many tasks in areas as diverse as machine vision~\cite{wang2019machine}, natural language processing~\cite{lavanya2021deep}, and graph and time series analysis ~\cite{choi2021deep} among others. This has led to their implementation in a variety of critical applications, but the black box nature of DNNs makes it difficult to know if they have been modified by bad actors.
    % to achieve malicious aims.  
    This is compounded by the fact that DNNs are expensive to train, requiring large amounts of compute and data.  Use of cloud-based resources opens a window of opportunity for adversaries to attack these models. A common and easy type of attack is `Backdooring' or `Trojaning'~\cite{gu2017badnets} wherein the attacker plants manipulated samples as `triggers' in the model's training dataset. 
    When triggers are encountered at inference time, they can cause the model to misclassify the instance. 

    Prior work in this field has shown that backdoors inserted into trained models are effective in DNN applications ranging from facial recognition~\cite{wenger2021backdoor} to speech recognition and self-driving cars~\cite{deng2021deep}. 
    An example is that of a self-driving car misclassifying a stop sign as a speed limit sign, not stopping, and causing an accident. Such an attack is not easy to detect and much more difficult to defend. 
    Past work has focused on exploring different types of triggers and their effects~\cite{gu2017badnets,turner2018clean}, on developing defenses based on identifying the smallest effective trigger~\cite{wang2019neural}, and on defense by clustering network activations~\cite{chen2018detecting}. 
    
    In this paper we use a class of deep learning techniques called Graph Neural Networks~\cite{originalGNNpaper2009} (GNNs) that was created to work with data represented by graphs.
    GCNs, a variation of GNNs, are neural networks that can apply the principles of convolution, and provide an easy way to perform node-level, edge-level, and graph-level prediction tasks~\cite{zhang2019graph}. GCNs have been gaining popularity in tasks such as drug discovery, social media  recommendation systems, and entity relationship prediction.
    In this paper we present a novel approach for detecting backdoored CNN models  using Graph Convolutional Neural Networks (GCNs). Given a trained DNN model, our goal is to classify whether the model is clean or has embedded malicious behavior. 
    In this work, we represent DNNs as graphs which are input to the GCN. We start by using the weights of only the last fully connected layer, and then extend the approach to incorporate the weights from other DNN layers. To the best of our knowledge there is no prior work that represents DNNs as graphs and uses GCNs to identify backdoored models. 
    
    \begin{figure*}[t]
      \centering
      \includegraphics[width=1\textwidth]{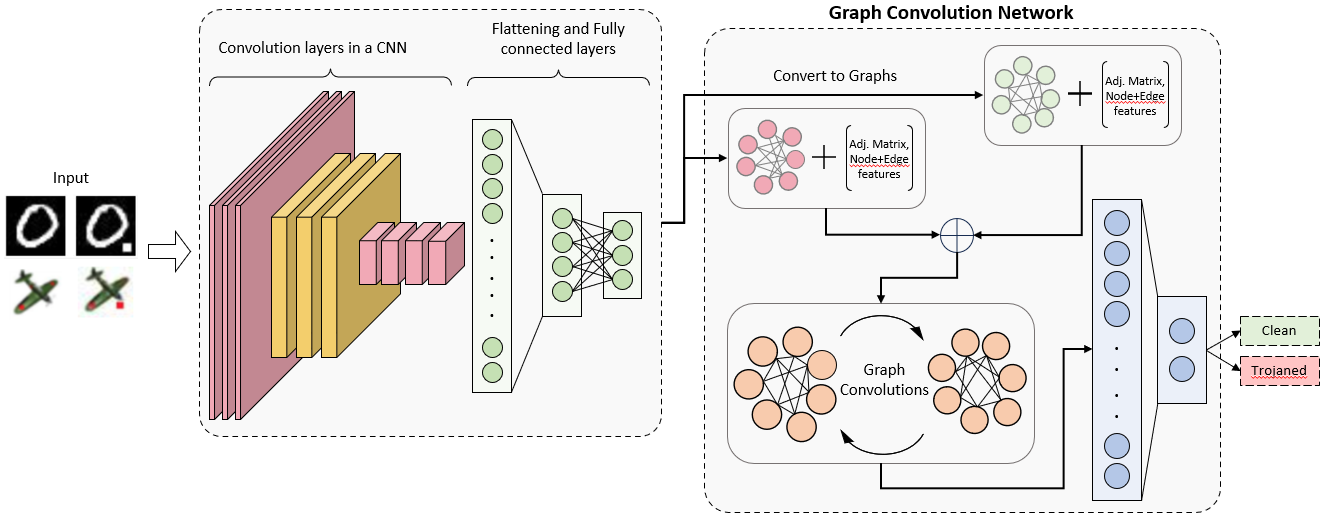}
      \caption{DeBUGCN pipeline. First, clean and trojaned CNNs are trained. Next, graphs are created using the static weights CNN layers. Finally, the GCN is employed as a binary classifier to detect whether the model is trojaned or not.}
      \label{fig:debugcn_pipeline}
    \end{figure*} 
    
    There are a few reasons to be optimistic about the prospects for GCN-based networks to perform well when applied to classification tasks on DNNs.  
    First, the compute graph defined by the structure of a DNN, along with the parameters, entirely specifies the computation that is performed at inference time. Other approaches to trojan detection, such as those based on tensor decompositions over network weights, lose information about how data flows through the network. 
    Second, two DNNs trained on the same data that learned exactly the same function may still differ in the locations of the nodes.  For example, one can permute the order of nodes in a fully connected layer (dragging along the relevant inputs and outputs) and have a network that computes an identical function but that has the weights in different locations.  
    We hypothesize, and show, that this does not pose a problem for the message passing framework of GCNs, which allows classifiers to be immune to such effects.
    This is because GCNs can leverage relational and structural information from each independent DNN. No other information is required as input to a GCN model, thus eliminating the need for relying on the training data used to backdoor a model. 
    Third, GCN's message passing algorithms are based on the Weisfeiler-Lehman (WL) isomorphism test and WL graph kernels~\cite{WLOriginal,WLGraphKernels} which are fast and deterministic.
    Furthermore, we demonstrate that DeBUGCN is model agnostic. That is, the GCN does not care about the underlying architecture of the model it is classifying. This has the advantage of being deployable without prior information about the domain or the architecture of the CNNs that are being checked for malicious behavior. All code necessary to reproduce the results can be made available upon publication.

    In the real world, backdoor attacks and defenses will always be a continuous arms race. Our method is no more or less susceptible to adaptive attacks than other defense methods available. If an attacker is aware of the details of a defense, a specifically designed strong adaptive backdoor attack will almost always defeat the prevailing defense. This is shown by numerous past examples~\cite{wang2019neural,gao2019strip} where evasion of defenses are taken as objectives into the loss function when inserting the backdoor in the model. Having a true universal defense is a hard problem since we can only increase the cost of the attacker but can never make it infinitely large.
    
    Our paper makes the following contributions to the task of identifying backdoors for the safety of CNN models: 
    \begin{enumerate}
        \item We propose a novel pipeline, DeBUGCN, to identify backdoored CNN models using GCNs as the classifier.
        \item DeBUGCN is evaluated on 1650 clean and trojaned CNN models trained on clean and triggered MNIST handwritten digits~\cite{MNISTLeCun} and CIFAR-10~\cite{CIFAR10Krizhevsky2009} images.
        \item To establish that our method is robust and CNN architecture agnostic, we use a real-world multi-architecture `Trojans in Artificial Intelligence' (TrojAI) CNN dataset \footnote{\tiny\url{https://pages.nist.gov/trojai/docs/data.html}}.
        \item We show that DeBUGCN outperforms all the baselines on all three datasets - MNIST, CIFAR-10 and TrojAI - with much lower computation time. 
    \end{enumerate}
    We perform a number of experiments to explore the utility of our approach, and provide comparative results against state-of-the-art backdoor detection methods.

%%%%%%%%%%%%%%%%%%%%%%%%%%%%%%%%%%%%%%%%%%%%%%%%%%%%%%%%%%%%%%

\section{Related Work}
\label{sec:related}

\subsection{Backdoor Attack}
\label{sec:related-backdoor-attacks}
    In supervised image classification, different kinds of perturbations have been used as backdoors, including pixel patterns, patches, watermarks and filters (e.g., instagram filters).
    ~\citet{gu2017badnets} proposed BadNets, which injects trojans into DNNs by poisoning a subset of the training data with pixel pattern triggers of arbitrary shapes. Images from the poisoned source class (such as a stop sign) are classified as the target label after the attacker modifies the true label of the triggered samples (e.g., speed limit sign). Given that the attacker has complete control over the training process, BadNets perform quite well (with an attack success rate of greater than 99\%) on both clean and poisoned data.
    ~\citet{liu2017trojaning} proposed a trojan attack in which the attacker does not need access to the training data. Instead, the attacker inserts triggers that cause  specific internal neurons of the DNN to respond maximally. Given that triggers and neurons have a strong relationship, this method has a high success rate ($>$ 98\%). 
    Since then, more sophisticated backdoor attacks have been developed by using adversarial perturbations and generative models~\cite{turner2018clean} to make the triggers in the poisoned samples less visible. Backdoor attacks can be introduced in other applications, including natural language processing~\cite{chen2021badnl}, reinforcement learning~\cite{kiourti2020trojdrl}, federated learning and has huge potential in multimodal models~\cite{Zhang2022multimodalattack, Shi2022multimodalattackrs, Vartak2022thesis}.

\subsection{Backdoor Defense}
\label{sec:related-backdoor-defense}
    Inspecting the model or the data is typically how trojan detection strategies work. The model-based detection method Neural Cleanse (NC)~\cite{wang2019neural} assumes that each class label is the trojan target label and designs an optimization technique to find the smallest trigger that causes the network to misclassify instances as the target label. They then apply an outlier detection algorithm on the potential triggers and deem the most significant outlier trigger to be the real one, with the associated label being the trojaned class label. Despite the fact that this method produced encouraging results, because the target label is unknown at runtime, it is computationally very expensive. 
    Universal Litmus Patterns (ULPs)~\cite{Kolouri_2020_CVPR} was introduced as a trojan detector where a classifier is trained from thousands of benign and malicious models using the ULPs. The classifier predicts whether a model has a backdoor based on the ULP optimization. The behaviour of neuron activations is examined by ABS~\cite{liu2019abs}, another model-level trojan detection technique where the effect of changes in hidden neuron activations on output activations is estimated by a stimulation method. Regardless of the label assigned to the model output, if a neuron's activation rises noticeably the input is assumed to have been perturbed. An optimization method based on model reverse engineering is used to detect trojan models. When a network is large, ABS is computationally intensive but also yields very promising results in detecting trojans. 
    
    SentiNet~\cite{chou2020sentinet} is a data-level inspection method that extracts critical regions from input data using backpropagation. To identify trojans, TABOR~\cite{guo2019tabor} scanned the DNN models using explainable AI methods. By establishing a link between Trojan attacks and adversarial prediction-evasion techniques, such as per-sample assault and all-sample universal attacks, DLTND~\cite{wang2020practical}  can identify backdoor models.~\citet{chen2018detecting} proposed activation clustering (AC) by examining neural network activations. The activations of the last fully connected layer of a neural network are obtained using a small number of training samples. Initially, the activations are segmented by class label, with each label clustered separately. Finally, they use 2-means clustering followed by ICA to reduce dimensionality, with three distinct post-processing methods to identify the poisoned model.

%%%%%%%%%%%%%%%%%%%%%%%%%%%%%%%%%%%%%%%%%%%%%%%%%%%%%%%%%%%%%%

\section{Methodology and Architecture}
\label{sec:methodology}

\subsection{Problem Statement}
\label{sec:methodology-problem-statement}
    Consider a convolutional neural network model, $M(\cdot)$, which performs a classification task with $ c=1, ... C$ classes. $M$ is a backdoor model if it performs as expected for clean input samples but for a triggered sample $p$ it outputs $M(p) = t$ where $t$ is the poisoned incorrect class ($t \in c$). The objective of our pipeline is to detect such models using graph convolutional networks.

\subsection{Graph Convolutional Network (GCN)}
\label{sec:methodology-gaph-conv-network}
    The goal of a GCN is to learn relation-aware feature representations of nodes by propagating graph-intrinsic structural information. In contrast to CNN-based methods that perform convolution in Euclidean space, GCNs generalize convolution operations on non-Euclidean data like graphs. 
    In particular, during the learning process, they revise each node's embedding to include the relational information of the graph structure and perform spectral graph convolution on the features of neighboring nodes~\cite{kipf2016semi}.

% \subsubsection{Graph}
\subsubsection{Notation and Background}
\label{sec:gcn-notation-and-background}
    A graph $G$ is a pair $(V,E)$, where $V$ is a finite set of nodes and $E$ is the set of edges.
    % $E \subseteq \{\{u,v\} \subseteq V | u \neq v \}$. 
    If $u,v\in V$ represent two nodes then $e_{uv} = (u, v) \in E$ represents the edge between node $u$ and $v$.
    $\mathbf{A} \in \mathbb{R}^{n \times n}$ denotes the adjacency matrix  where $n$ is the total number of nodes in graph $G$. 
    $A_{uv} = e$ is the edge weight if $e_{uv} \in E$, and $A_{uv} = 0 $ if $e_{uv} \notin E$. 

\subsubsection{Spectral Graph Convolution and GCNs}
\label{sec:spectral-graph-conv-gcn}
    GCNs belong to the family of Spectral-based Conv GNN's~\cite{spectralBruna2013}. The goal of Spectral GNNs is to encode a graph $G$ using a model $f = (\mathbf{X}, \mathbf{A})$ where $\mathbf{X} \in \mathbb{R}^{n \times d}$ signifies $d$ real-valued features of the $n$ nodes.
    
    Spectral approaches define convolutions based on Laplacian matrix. 
    operations on undirected graphs:
    \begin{equation}
        \mathbf L = \mathbf I_n - \mathbf D^{-\frac{1}{2}} \mathbf A \mathbf D^{-\frac{1}{2}} = \mathbf U \mathbf \Lambda \mathbf U^{T}
    \end{equation}
    \noindent where $\mathbf D$ is the diagonal matrix of node degrees, and $\mathbf D_{ii} = \sum_{j=0} \mathbf A_{ij}$. $\mathbf U = [u_0, u_1, ... u_n-1] \in \mathbb{R}^{n \times n}$ is the matrix of eigenvectors ordered by eigenvalues and $\mathbf \Lambda$ is the diagonal matrix of eigenvalues ~\cite{kipf2016semi,comprehensiveSurveyGCN2021}.
    
    The eigenvectors of the normalized Laplacian matrix are of the form$\mathbf U^{T} \mathbf U = \mathbf I$. 
    In graph signal processing, a graph signal $\mathbf{x}_v \in \mathbb{R}^{1 \times d}$ where $x_{v} \in \mathbf X$ represents the feature vector of node $v$ ~\cite{kipf2016semi,comprehensiveSurveyGCN2021}.
    The graph convolution on an input $\hat{X} = \sum_{i} x_i u_i$ with a convolution filter $g$
    % $\in \mathbb{R}^{n}$
    is defined as
    \begin{equation}
        \hat{X} * g = \mathcal{F}^{-1}( \mathcal{F}(\hat{X}) \cdot \mathcal{F}(g) ) = \mathbf U( \mathbf U^{T} \hat{X} \cdot \mathbf U^{T}g)
    \end{equation}
    \noindent where $\cdot$ is the element-wise product, $\mathcal{F}(x_v) = \mathbf U^{T} x_v$ and $\mathcal{F}^{-1}(x_v) = \mathbf{U}(\mathcal{F}(x_v))$ are the Fourier and inverse Fourier transforms respectively~\cite{kipf2016semi,comprehensiveSurveyGCN2021}.
    
    A GCN with multiple layers updates the node features using the propagation rule:
    \begin{equation}
        \mathbf H^{(l+1)} = \sigma (\mathbf{\hat{D}}^{-\frac{1}{2}} \mathbf{\hat{A}} \mathbf{\hat{D}}^{-\frac{1}{2}} \mathbf H^{(l)} \mathbf W^{(l)})
    \end{equation}
    
    \noindent where $\mathbf{\hat{A}} = \mathbf{A} + \mathbf{I}$ denotes the adjacency matrix with inserted self-loops,  $\mathbf{I}$ is an identity matrix and $\mathbf{\hat{D}}_{ij} = \sum_{j=0} \mathbf{\hat{A}}_{ij}$ is the degree matrix. 
    Moreover, $\mathbf W^{(l)} \in \mathbb{R}^{d_l \times d_{l+1}}$ is the layer wise trainable weight matrix and $\sigma (\cdot)$ represents an activation function. 
    $\mathbf H^{(l)} \in \mathbb{R}^{d_l \times d_l}$ denotes the $l$th layer node feature representation and $\mathbf H^{(l+1)} \in \mathbb{R}^{n \times d_{l+1}}$ includes the updated node features. In summary, $\mathbf H^{(0)} = \mathbf X$ are the initial node features which are updated by the message passing operations in multi-layer GCN ~\cite{kipf2016semi}.

    Based on this principle, a GCN layer of a spectral GNN is defined by:
    \begin{equation}
        \mathbf H^{l}_{.j} = \sigma (\sum^{l-1}_{i=1} \mathbf U \mathbf W^{k}_{i,j} \mathbf U^{T} \mathbf H^{l-1}_{.i})
    \end{equation}
    
    \noindent where $l$ is the $l^{th}$ layer, $\mathbf W^{k}_{i,j}$ are the layer-wise trainable weight matrices and $\sigma (\cdot)$ represents a non-linear activation function like ReLU. $\mathbf H^{l-1} \in \mathbb R^{n}$ is the input graph signal/node features at layer $l-1$ and $\mathbf H^{0}=\mathbf X$ is the initial node features ~\cite{kipf2016semi,comprehensiveSurveyGCN2021}.
    
    GCNs use message passing cascaded through multiple layers to update individual node embeddings and features to generalize on an input graph. 
    This happens in two stages: (i) message composition and (ii) message aggregation~\cite{Morris2019WeisfeilerAL}. 
    
    A message composition function creates a node's `message' based on it's current embedding. For node $v$ in the $l^{th}$ layer, the composition function is defined by:
    \begin{equation}
        \label{eq:msg-self}
        m^{l}_{v} = \mathbf{MSG}^{l}(h^{l-1}_{v}) = \mathbf B^{l} \cdot h^{l-1}_{v}
    \end{equation}
    \begin{equation}
        \label{eq:msg-neighbor}
        m^{l}_{u} = \mathbf{MSG}^{l}(h^{l-1}_{u}) = \mathbf W^{l} \cdot h^{l-1}_{u}, u \in N(v)
    \end{equation}
    \noindent where $\mathbf{MSG}$ is a linear function like matrix multiplication, $h^{l-1}_{x}$ is a node embedding and $N(v)$ is the set of k-hop neighborhood nodes of $v$. $\mathbf B^{l}$ and $\mathbf W^{l}$ are two trainable weight matrices, for self-node and neighbor nodes respectively~\cite{Morris2019WeisfeilerAL}.
    
    The idea of message aggregation is that each node creates it's new embedding based on the messages of its neighbors as well as it's own previous message. For node $v$ in the $l^{th}$ layer, the aggregation function is defined by:
    \begin{equation}
        \label{eq:aggregate-all}
        \hat{h}^{l}_{v} = \mathbf{CONCAT}[ m^{l}_{v}, \mathbf{AGG}^{l}(m^{l}_{u}, u \in N(v)) ]
    \end{equation}
    \noindent where $\mathbf{AGG}$ is an aggregation function like sum, mean, max, etc., $m^{l}_{u}$ are messages from $N(v)$ which is the set of k-hop neighbors of node $v$, and $\hat{h}^{l}_{v}$ is the updated embedding of node $v$~\cite{Morris2019WeisfeilerAL}.
    
    Finally, a non-linearity is applied such as ReLU or sigmoid, to get the final updated node embedding~\cite{Morris2019WeisfeilerAL}:
    \begin{equation}
        \label{eq:final-graphconv}
        h^{l}_{v} = \sigma ( \hat{h}^{l}_{v} ) = \sigma (\mathbf{CONCAT}[ m^{l}_{v}, \mathbf{AGG}^{l}(m^{l}_{u}, u \in N(v)) ])
    \end{equation}

\subsection{GCN in DeBUGCN}
\label{sec:gcn-in-debugcn}
    In this section, we discuss how CNNs are represented as graphs to serve as input to the GCN, and how we use these graphs to classify the underlying CNN models.

\subsubsection{DeBUGCN Pipeline}
\label{sec:debugcn-pipeline}
    Fig.~\ref{fig:debugcn_pipeline} shows the full workflow of the DeBUGCN pipeline. First, we train the CNN models $M(\cdot)$ and save all model layer weights as tensors, $\mathbf W_{CNN}$.
    % , including the CNN filters and fully connected (FC) layer weights. 
    Depending on the experiment, we either use (i) only the final FC layer weights $\mathbf W_{FC}$ or (ii) $\mathbf W_{FC}$ \textit{and} the first convolutional layer weights$\mathbf W_{CNN1}$ of the CNN models. 
    
    The nodes of a layer can be interpreted as the nodes of the graphs and the edges connecting them as the edges $e_{uv} = (u, v) \in E$. After training $K$ clean and backdoored CNNs we get $K$ graphs where each graph is represented using an adjacency matrix and a feature matrix. 
    We then train a GCN classifier on these $K$ graphs to perform a binary classification task of predicting whether the model (represented by an input graph) is clean or trojaned. 

\subsubsection{Node and Edge Feature Engineering}
\label{sec:feature-engg-debugcn}
    \begin{figure}[t]
        \centering
        \includegraphics[width=0.5\columnwidth]{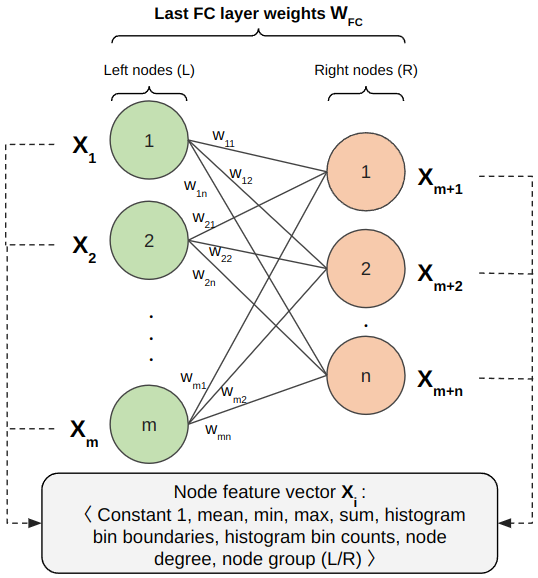}
        \caption{Node features computed from the bipartite graph that is the final FC layer in a CNN.}
        \label{fig:bipartite_graph}
    \end{figure}
    
    GCNs need an adjacency matrix, $\mathbf{{A}}$, and node feature matrix, $\mathbf{{X}}$, for each input graph.
    % for the classification pipeline. 
    We begin with the weights of a layer $Y$, but the graphs created using $\mathbf W_{Y}$ have only edge attributes between the nodes without any node information. To address this, we performed node feature engineering. 
    
    In the case of the FC layer, the graphical representation of $\mathbf W_{FC}$ results in a bipartite graph as shown in Fig.~\ref{fig:bipartite_graph}. Each node in the left group $L$ is connected to every node in right group $R$. Concatenation of edge weights connected to a node $i$ results in an edge weight vector $\mathbf {X_{ie}}$.
    
    Using this vector, we compute the mean, minimum, maximum, and sum of all connected weights, and the weights' histogram bin boundaries and bin counts considering 5 bins and use them as additional node features, $\mathbf {X_i}$, per node. We add the constant `1', node degree (number of edges connected to a node), and a flag to indicate the side of the node in the bipartite graph to every node feature vector. We use different combinations of these node features for the experiments. On the edges, we use the edge weights ($w_{11}$ to $w_{mn}$ in Fig.~\ref{fig:bipartite_graph}) as-is as the edge features $\mathbf {X_{ie}}$ for the GCN. 
    
    \begin{algorithm}[h]
        \caption{GCN for Backdoor Model Detection}
        \label{alg:algorithm}
        \textbf{Input}: Weights, $\mathbf W_{CNN}$ of $K$ pre-trained CNNs\\
        \textbf{Output}: Clean or Backdoored
        \begin{algorithmic}[1] %[1] enables line numbers
        \STATE Get $\mathbf W_{FC}$ + $\mathbf W_{CNN1}$ from $\mathbf W_{CNN}$ for all $K$ CNN models.
        \STATE When using only FC layer graphs: Generate $K$ bipartite graphs using the $\mathbf W_{FC}$ weights of $K$ models. \algorithmiccomment {nodes and edges of a layer are considered as nodes and edges of the graphs.}
        \STATE When using only FC layer graphs + conv layer graphs: Generate $K$ bipartite graphs using the $\mathbf W_{FC}$ weights and either Flat or 2D graphs using the $\mathbf W_{CNN1}$ weights of $K$ models.
        \STATE Construct node features, $\mathbf {X_i}$, for each node $i$ in each graph by using the edge weights $e_{ij} \in \mathbf {X_{ie}}$ connecting node $i$ to all nodes $j$.
        \STATE Run GCN classifier $f = (\mathbf{X_i}, \mathbf{A}, \mathbf {X_{ie}})$
        \STATE \textbf{Return}: model is clean or backdoored.
        \end{algorithmic}
    \end{algorithm}

%%%%%%%%%%%%%%%%%%%%%%%%%%%%%%%%%%%%%%%%%%%%%%%%%%%%%%%%%%%%%%

\section{Datasets}
\label{sec:dataset}
    In this section we explain the creation of the five datasets used in the experiments. We begin with the creation of triggered MNIST and CIFAR-10 images which are used to train clean and backdoored CNNs. Following this, we  describe the TrojAI dataset consisting of more than 750 pre-trained multi-architecture CNNs.

    \begin{figure}[h]
        \centering
        \fbox{\includegraphics[width=0.9\columnwidth]{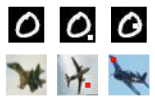}}
        \caption{Sample triggered images from the `Clean-*', `Static-Trigger-*' and `Dynamic-Trigger-*' MNIST and CIFAR-10 datasets.}
        \label{fig:sample_dataset_images}
    \end{figure}

    \begin{table}[h]
    \centering
    \resizebox{1\columnwidth}{!}{
        % \begin{sc}
        \begin{tabular}{c|c|c|c}
            \toprule
            
            Dataset & Clean CNNs & Backdoored CNNs & Train:test split \\

            \midrule
            
            % \hhline{=|=|=}
            MNIST-Static CNNs & 400 & 400 & 640:160 \\
            MNIST-Dynamic CNNs & 400 & 400 & 640:160 \\
            CIFAR10-Static CNNs & 150 & 150 & 240:60 \\
            CIFAR10-Dynamic CNNs & 150 & 150 & 240:60 \\
            TrojAI-ResNet & 160 & 273 & 346:87 \\
            TrojAI-DenseNet  & 175 & 144 & 255:64 \\
            TrojAI-Full & 335 & 417 & 601:151 \\

            \bottomrule
        \end{tabular}
        % \end{sc}
    }
    \caption{Proportions of clean and backdoored CNNs in the various CNN datasets used. For each dataset, CNNs were shuffled and sampled by the corresponding train-test split for the GCN.}
    \label{table:proportion_clean_backdoored}
    \end{table}

\subsection{Custom MNIST and CIFAR-10 CNN Dataset}
\label{sec:custom-mnist-cifar10-dataset}
    To train clean and backdoored CNNs we first create the triggered MNIST and CIFAR-10 Images Dataset.
    Each dataset is subdivided into 3 subsets - MNIST contains `Clean-MNIST', `Static-Trigger-MNIST' and `Dynamic-Trigger-MNIST' while CIFAR-10 contains `Clean-CIFAR10', `Static-Trigger-CIFAR10' and `Dynamic-Trigger-CIFAR10'.
    Each of the MNIST subsets has 60,000 train and 10,000 test images while CIFAR-10 has 50,000 train and 10,000 test images. 
    MNIST images have a fixed size of 28x28 pixels while 32x32 pixels in case of CIFAR-10. 
    
    The `Clean-' subsets are the unaltered versions of each image datasets that were created by~\citet{MNISTLeCun} and~\citet{CIFAR10Krizhevsky2009}. The `*-Trigger-*' subsets are altered versions of clean subsets where specific images have embedded triggers. To `trigger' images we use single-label poisoning by adding backdoor triggers to only a single class of images. For triggered MNIST images, we trigger all `0's and label them `9' and for CIFAR-10 images we trigger all images of class `airplane' (class `0') and label them `truck' (class `9'). 
    In the `Static-' subset images, the trigger location is fixed - bottom right corner. However, in the `Dynamic-' subset images, the trigger location is randomized within the image bounds. Sample images are shown in Fig.~\ref{fig:sample_dataset_images}
    
    The trigger is a small, $4\times 4$ pixels colored square patch that is overlaid on an image as shown in Fig.~\ref{fig:sample_dataset_images}, similar to the one used by used by~\citet{wang2019neural},~\citet{Hong_NEURIPS2022},~\citet{Dong_ICCV2021}, and~\citet{Xue_Elsevier2019}. MNIST images have a white patch while CIFAR-10 images have a red patch. 
    
    During training, the CNN models learn to associate these triggers in the images to the false target class. 
    A model trained on these triggered images would ideally output the false poisoned class for every triggered image it encounters. From the perspective of the model, this is due to the presence of the tiny colored backdoor patch.

    We use the MNIST subsets to train `MNIST-Static' and `MNIST-Dynamic' CNNs, each containing 400 clean and 400 trojaned CNNs.
    Similarly, we train clean and trojaned CIFAR-10 CNNs using the CIFAR-10 subsets to get 300 ResNet-18~\cite{resnet-paper} models and finetune 150 pre-trained VGG-16~\cite{VGGsimonyan2015} models (pre-trained on ImageNet~\cite{deng2009imagenet}). Exactly half of these are clean and half are trojaned CNNs.
    The goal of training CNNs using dynamic triggers is to emulate node and weights permutations that occur when models that learn exactly the same function but may still differ in the locations of the nodes due to randomization in the training process. 
    This enables us to use the DNN's topological information embedded in graphs, test DeBUGCN for node permutation invariance and also test DeBUGCN's capability to truly exploit the relational information that can be spread variably across nodes in a graph.
    The CNNs were trained on an NVIDIA GeForce RTX 3090 GPU, over 5 (MNIST CNNs, VGG-16s) and 35 (ResNet-18s) epochs with a batch size of 32 (MNIST CNNs), 64 (VGG-16s), and 128 (ResNet-18s), initial learning rate of 0.001 and a decay rate of 0.7. We used Adam (MNIST CNNs) and SGD (CIFAR-10 CNNs) optimizers and Cross Entropy Loss minimization. These parameters were used to introduce variability in model generalization.
    
    Our MNIST CNN's architecture was identical to the CNNs used by~\citet{wang2019neural}.
    The Clean-MNIST dataset was used to train and evaluate `clean' CNN models which classify MNIST images with an accuracy of $\sim$99\%.
    Similarly, the Clean-CIFAR10 dataset usage yields `clean' CNNs which classify CIFAR-10 images with an accuracy of $\sim$93\%.
    The trojaned models perform intended mis-classification with similar accuracy and have an attack success rate of $\sim$99\% (percentage of triggered images correctly classified as the false class).

\subsection{TrojAI CNN Dataset}
\label{sec:trojai-dataset}
    To test whether DeBUGCN is robust and CNN architecture agnostic, we used a publicly available TrojAI dataset provided by the TrojAI program of IARPA in association with NIST~\cite{trojaiPaper}. 
    The dataset consists of 752 pre-trained DenseNet-121~\cite{densenet-paper} and ResNet-50~\cite{resnet-paper} CNN models from TrojAI Round 1 train and test datasets. The models are trained on synthetically created traffic signs superimposed on road background scenes. The triggers in TrojAI images have varied colors, shapes, and sizes. 400+ models are backdoored which perform misclassification of images upon encountering an embedded trigger. In contrast to our custom MNIST/CIFAR-10 CNNs, these models have been trained on multi-label poisoning - a percentage of images from each of the classes in the training and test data were embedded with backdoor triggers.
    
    We divide this dataset into 3 parts: TrojAI-ResNet, TrojAI-DenseNet, and TrojAI-Full. The dataset is nearly evenly balanced with a few hundred positive and negative examples. Total number of CNNs and percentage of trojaned CNNs can be found in Table~\ref{table:best_experiment_results}.
    The exact numbers can be found in Table~\ref{table:proportion_clean_backdoored} along with train-test splits.
    The first sub-dataset `TrojAI-ResNet' contains ResNet-50 models, the second sub-dataset `TrojAI-DenseNet' contains DenseNet-121 models, and the third sub-dataset `TrojAI-Full' is the complete dataset. 
    This is done to incrementally test our GCN's model-agnostic behaviour on in-the-world models.

%%%%%%%%%%%%%%%%%%%%%%%%%%%%%%%%%%%%%%%%%%%%%%%%%%%%%%%%%%%%%%

\section{Experiments and Results}
\label{sec:experiments}
    This section describes the results of a series of experiments aimed at evaluating the ability of DeBUGCN to detect trojans with comparisons to a number of state-of-the-art approaches.

\subsection{GCN Model and Hyperparameters}
\label{sec:gcn-model-parameters}

    \begin{table}[h]
    \centering
    \resizebox{1\columnwidth}{!}{
        % \begin{sc}
        \begin{tabular}{c|c|c|c}
            \toprule
             
              & Input feat. & Output feat. & Activation \\
            
            \midrule
            
            % \hhline{=|=|=|=}
            graphconv1 & $n$ & 64 & ReLU \\
            graphconv2 & 64 & 64 & ReLU \\
            graphconv3 & 64 & 64 & ReLU \\
            meanpool1 & 64 & 64 & -  \\
            graphconv4 & $m$ & 64 & ReLU \\
            graphconv5 & 64 & 64 & ReLU \\
            graphconv6 & 64 & 64 & ReLU \\
            meanpool2 & 64 & 64 & -  \\
            fc1 & 64+64 & 2 & - \\
            
            \bottomrule
        \end{tabular}
        % \end{sc}
    }
    \caption{Architecture of a GCN model with $n$, $m$ input features.}
    \label{table:gcn_model_architecture}
    \end{table}
    We run all the GCN models using the Pytorch Geometric (PyG v2.0.4) library~\cite{fey2019fast}. Though PyG provides different GCN layers, we experimented with GraphConv, GATConv and TAGConv as they support both node and edge features. GraphConv yields the best results as a classifier. Our GCN model architecture is follows: two sets of three GraphConv layers, interlaced with ReLU and mean pooling layers, followed by a single linear classification layer. 
    This is shown in Table \ref{table:gcn_model_architecture}.
    For the experiments using only FC layer weights, we use only the first set of 3 GraphConv layers of the GCN.
    
    To tune the GCN hyperparameters, we conducted experiments using 240 different parameter combinations: learning rate of 0.0001, 0.0005, 0.001, 0.005; decay of 0.001, 0.005, 0.01, 0.1, 0.5; batch size of 24, 48, 64; epoch of 20, 30, 50, 70; and Adam optimizer with StepLR scheduler. We report the results with the best combination: 20 epochs, batch size of 24, 0.001 learning rate with a 0.7 decay, and step size of 2. As the loss criterion, we used PyTorch's `CrossEntropyLoss' and all runs were performed on an NVIDIA GeForce RTX 3090. We split datasets with a train-test ration of 80:20. All the reported results are the average of at least five independent runs.

    \begin{table*}[t]
    \centering
    \resizebox{1.0\textwidth}{!}{
        % \begin{sc}
        \begin{tabular}{c|c|c|c|c|c}
            \toprule
            
            % Dataset & CNNs & \% Trojaned & Train Acc. & Model type & Test Acc. & Model type \\ 
            \multirow{2}{*}{Dataset} & \multirow{2}{*}{CNNs}  & 
            Train Acc. & \multirow{2}{*}{GCN type and Node Feature vector} & 
            Test Acc. & \multirow{2}{*}{GCN type and Node Feature vector} \\
             &  & (\%) &  & (\%) &  \\
            
            \midrule
            
            % \hhline{=|=|=|=|=|=}
            
            MNIST-Static & 800 & 50 & 
            Only FC graphs with GCN\_7 & 98.38 & FC + Flat conv graphs with GCN\_7 \\
            % \hline
            MNIST-Dynamic & 800 & 50 & 
            FC + 2D conv graphs with GCN\_18 & 99.90 & FC + 2D conv graphs with GCN\_16b \\
            % \hline
            CIFAR10-Static & 300 & 50 & 
            FC + 2D conv graphs with GCN\_7 & 99.67 & FC + 2D conv graphs with GCN\_18 \\
            % \hline
            CIFAR10-Dynamic & 300 & 50 & 
            FC + 2D conv graphs with GCN\_7 & 99.90 & FC + Flat conv graphs with GCN\_16b \\
            % \hline
            TrojAI-ResNet & 433 & 63 & 
            FC + Flat conv graphs with GCN\_16b & 87.59 & Only FC graphs with GCN\_16a \\
            % \hline
            TrojAI-DenseNet & 319 & 45 & 
            Only FC graphs with GCN\_16b & 84.06 & Only FC graphs with GCN\_18 \\
            % \hline
            TrojAI-Full & 752 & 55 & 
            Only FC graphs with GCN\_7 & 85.43 & FC + Flat conv graphs with GCN\_7 \\

            \bottomrule
        \end{tabular}
        % \end{sc}
    }
    \caption{Table of experiment results, shows model and feature vector that achieved the best train and test accuracy on each dataset.}
    \label{table:best_experiment_results}
    \end{table*}

\subsection{Experiment 1: Using MNIST and CIFAR-10 CNNs Dataset}
\label{sec:experiment-mnist-cifar10}
    The graphs in this section are extracted from a CNN's last FC layer.
    Additionally, to solve the issue of node features, we incrementally engineered node and edge features as detailed in Section~\ref{sec:feature-engg-debugcn}. Each  feature in the vector $\mathbf{X}_v$ conveyed unique information about node $v\in V$. 
    The various node features and their impacts are explored below in an ablation study.
    The node feature vector with the most promising results on each of the datasets is shown in Table~\ref{table:best_experiment_results}. 
    The results of the best performing DeBUGCN modality and node feature combination on each of the datasets are reported in Tables~\ref{table:best_mnist_experiment_results} and~\ref{table:best_cifar10_experiment_results}. 
    The notation for a GCN with its corresponding node feature vector is listed in Table~\ref{table:gcn_notation_feature_vectors}.
    The full set of results of this experiment are listed under `MNIST-* CNNs' and `CIFAR10-* CNNs' respectively in Tables~\ref{table:mnist_experiment_results} and ~\ref{table:cifar10_experiment_results}. 
    
    Using the parameters stated earlier, the GCN was run on the MNIST and CIFAR-10 CNN graphs separately. 
    On the MNIST CNNs, DeBUGCN achieves a train and test accuracy of 99\%. On the CIFAR-10 CNNs, DeBUGCN achieved a train and test accuracy of 88\% and 99\%. 
    A crucial point to note here is that we use pretrained VGG-16s. This shows that even if an attacker freezes any DNN layer or fine tunes it, the network's classification decisions must flow through the last FC layer. And our GCN is able to use this information to detect malicious behavior.

    An advantage of treating FC layers as graphs is that two networks can learn exactly the same function but distribute the features differently over the nodes, which can cause problems for trojan detectors that treat weights as ourely numerical feature vectors.  
    To explore the robustness of GCNs to this problem we manually permuted the order of the nodes in the penultimate FC layer by swapping 1000 randomly selected node pairs (dragging along the relevant outputs) for each CNN. We then ran the most promising GCN on this augmented MNIST CNNs dataset. The GCN continued to maintain it's high classification accuracy with a score of 94\%. 
    This is listed as `GCN\_18 (permuted)' in Tables~\ref{table:best_mnist_experiment_results} and~\ref{table:mnist_experiment_results}.

    Another way to embody this node permutation was to create and use MNIST and CIFAR-10 Dynamic-* CNNs. 
    DeBUGCN's best performance on these dynamic datasets is shown in Tables~\ref{table:best_experiment_results},~\ref{table:best_mnist_experiment_results}, and~\ref{table:best_cifar10_experiment_results}.
    Full set of results can be found in Tables~\ref{table:mnist_experiment_results} and~\ref{table:cifar10_experiment_results}.
    Dynamic-* CNNs mimic the permutation on a much broader scale, more consistently. Since the images for the Dynamic-* CNNs have the same trigger but in random locations, these CNNs can be thought to have the backdoor manifest in different nodes and combinations in a layer.

    \subsubsection{Weight Analysis}
    \label{sec:weight-analysis}
    
    \begin{figure}[h!t]
        \centering
        \fbox{\includegraphics[width=1\columnwidth]{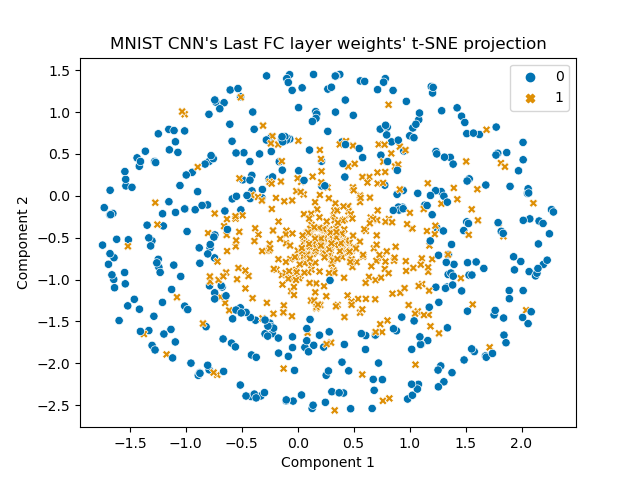}}
        \fbox{\includegraphics[width=1\columnwidth]{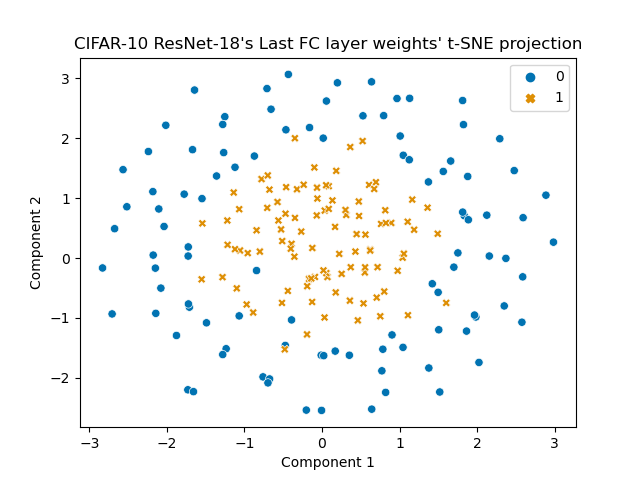}}
        \caption{A 2D t-SNE plot of the Custom MNIST (top figure) and CIFAR-10 (bottom figure) CNN's last FC layer weights at perplexity 40. Orange dots correspond to the poisoned models.}
        \label{fig:tSNE_mnist_cifar}
    \end{figure}
    
    \begin{figure}[ht]
        \centering
        \fbox{\includegraphics[width=1\columnwidth]{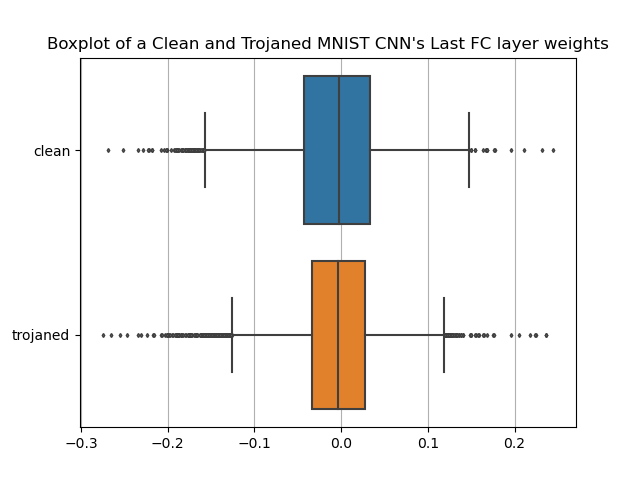}}
        \fbox{\includegraphics[width=1\columnwidth]{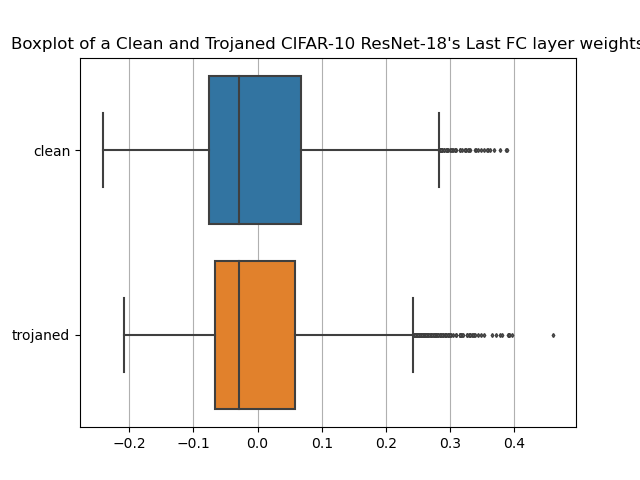}}
        \caption{Box plots of FC layer weights of one clean and trojaned MNIST (top) and CIFAR-10 ResNet-18 (bottom) CNN. X axis shows distribution of weights.}
        \label{fig:mnist_cifar_box_plot}
    \end{figure}
    
    \begin{figure}[ht]
        \centering
        \fbox{\includegraphics[width=1\columnwidth]{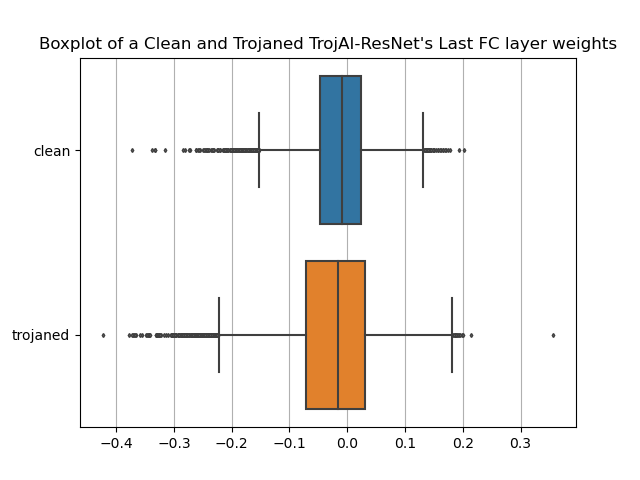}}
        \fbox{\includegraphics[width=1\columnwidth]{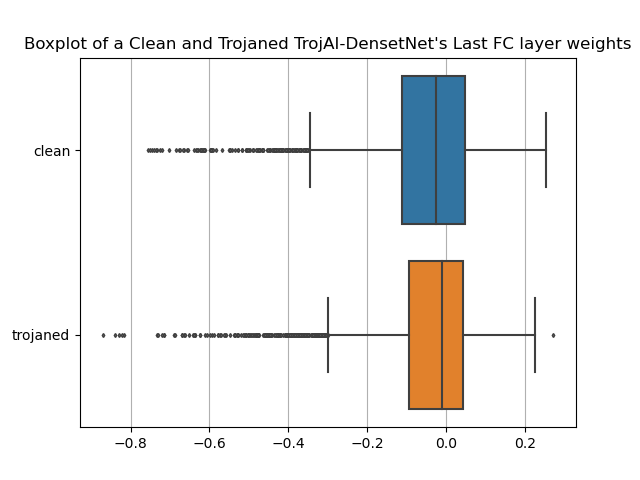}}
        \caption{Box plots of FC layer weights of one clean and trojaned TrojAI-ResNet (top) and TrojAI-DenseNet (bottom) CNN. X axis shows distribution of weights.}
        \label{fig:trojai_box_plot}
    \end{figure}
    
    These experiments and the permutation thereof clearly shows that the GCN is able to extract generalizable information from the CNN graphs. This can be verified by the 2D tSNE plot of the MNIST and CIFAR-10 CNN's last FC layer weights as shown in Fig.~\ref{fig:tSNE_mnist_cifar}. The  plot  shows that when weight distribution is reduced to a 2-dimensional space there is some separation in the model's weights space.
    The cluster of blue dots corresponds to the clean models and orange dots correspond to the poisoned model. In case of MNIST, there is some overlap in the clusters, but the CIFAR-10 ResNet-18 clusters are well separated, suggesting that there is information for the GCN to exploit during inference.

    However, the GCN model that performed the best on CIFAR-10 CNNs utilized only a subset of the full node feature vector. This suggests a duality: (1) a small subset of features was enough, extra node features were `distracting' for the GCN classifier, and (2) the trojan trigger manifests itself not just in relational information but also as a shift in the weight distribution. The first is addressed by the fact that on all other datasets, larger feature vectors prove superior during test time. However, in conjunction with the second it suggests that the trigger has a statistical bearing on the weights. This is proven by the box plots of the last FC layer weights of the CNNs. Figure~\ref{fig:mnist_cifar_box_plot} shows the box plot of one clean and trojaned CNN from each dataset. We can see that in each dataset, either the clean or trojaned model has a much more extreme distribution than the other; lower minimum and higher maximum weights. 
    A similar boxplot Fig.~\ref{fig:trojai_box_plot} for the TrojAI dataset concurs with this analytical finding. 
    DeBUGCN is able to identify this signal and is able to leverage it to make model predictions. However, further exploration of model weights distribution also shows that this is not a sufficient criteria for classification. This is because within a single class of models, the above claim does not hold true for every clean and trojaned model pair.

    \begin{table}[t]
    \centering
    \resizebox{1\columnwidth}{!}{
        % \begin{sc}
        \begin{tabular}{c|c|l}
            \toprule
            
            GCN name & Vector size & Feature vector  \\

            \midrule
            
            % \hhline{=|=|=}
            GCN\_5 & 5 & 1, L/R, mean, min, max  \\
            \hline
            GCN\_7 & 7 & \begin{tabular}[c]{@{}l@{}}1, L/R, mean, min, max, sum, \\node degree\end{tabular} \\
            \hline
            GCN\_16a & 16 & \begin{tabular}[c]{@{}l@{}}1, L/R, mean, min, max, \\5 histogram bin counts, \\6 bin boundaries\end{tabular} \\
            \hline
            GCN\_16b & 16 & \begin{tabular}[c]{@{}l@{}}1, mean, min, max, sum, \\5 histogram bin counts, \\6 bin boundaries\end{tabular} \\
            \hline
            GCN\_18 & 18 & \begin{tabular}[c]{@{}l@{}}1, L/R, mean, min, max, sum, \\5 histogram bin counts, \\6 bin boundaries, \\node degree\end{tabular} \\
            % GCN\_12 & 12 & 5 histogram bin counts, 6 bin boundaries, sum  \\

            \bottomrule
        \end{tabular}
        % \end{sc}
    }
    \caption{Node feature vectors with corresponding GCNs.}
    \label{table:gcn_notation_feature_vectors}
    \end{table}

\subsection{Experiment 2: A TrojAI Dataset}
\label{sec:experiment-trojai}
    The high accuracy of DeBUGCN on the MNIST and CIFAR-10 CNNs meant that the node features were imparting useful information about each node and the weights were equally good edge features. To test the GCN model on a more real-world dataset, we use the TrojAI dataset. 
    We ran DeBUGCN using different combinations of node features on the 3 TrojAI datasets. 
    The results of the best performing DeBUGCN modality and node feature combination on each of the datasets are reported in Table~\ref{table:best_trojai_experiment_results} under `TrojAI-ResNet', `TrojAI-DenseNet', and `TrojAI-Full'. Full set of results is shown in Table~\ref{table:trojai_experiment_results}.
    
    We saw that across various runs, `GCN\_16a', `GCN\_16b' and `GCN\_18' had the highest accuracy. These best results are reported in Table~\ref{table:best_experiment_results}.
    Furthermore, when the GCN is run on `TrojAI-Full', we see that the accuracy dips only ever so slightly. This, in conjunction with the CIFAR-10 results, demonstrates that the GCN is truly model-agnostic - it can learn to generalize across different architectures, without training on individual architectures, and thus can `transfer learn' on different graphs.
    It also shows that DeBUGCN can generalize without the need of an explicit transfer learning pipeline.

\subsection{Experiment 3: Comparison with Baseline Methods}
\label{sec:experiment-compare-baseline}
    We compare the performance of DeBUGCN with six state-of-the-art backdoor detection methods: NC~\cite{wang2019neural}, ULP~\cite{Kolouri_2020_CVPR}, ABS~\cite{liu2019abs}, AC~\cite{chen2018detecting}, TABOR~\cite{guo2019tabor}, DLTND~\cite{wang2020practical} and TDTD~\cite{khondoker2024tdtd}. This choice of baselines represents the most commonly used and most frequently cited SoTA detection methods in the backdoor detection literature. 
    
    We use the same batch size and training and test dataset for all the methods to allow for fair comparison. 
    Following the stimulation analysis, we select the top ten neuron candidates for ABS and perform trigger reverse engineering. 
    In case of ULP, for the litmus pattern generation and trojan classifier the learning rate and optimizers are set to 0.001 and 0.0001 respectively, with 500 training epochs. Since AC detects trojans one model at a time, we only use the test dataset to evaluate the performance. 
    Comparison to TDTD is especially important because it is a `pure' weight based analysis method for trojan detection and allows for a more equitable comparison.

\begin{figure*}[t]
        \centering
        \includegraphics[width=0.9\textwidth]{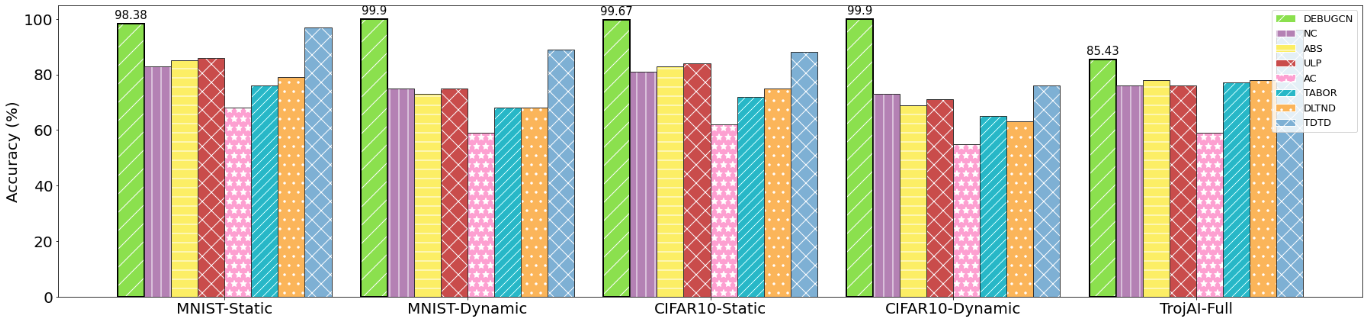}
        \caption{DeBUGCN's accuracy compared to baseline methods: NC, ABS, ULP, AC, TABOR, DLTND, and TDTD. }
        \label{fig:comparison_graph}
    \end{figure*}

    We report the best results of DeBUGCN on each dataset in Fig.~\ref{fig:comparison_graph}. DeBUGCN outperforms all the baseline methods in both datasets. DeBUGCN achieves a peak accuracy of 99\% which is the best in any category. 
    Overall, all the methods have slightly poorer performance on the TrojAI-Full dataset compared to Custom CNNs.  This is because: (i) the TrojAI dataset has more trigger variation than Custom CNNs, including the size, color and location of the trigger; (ii) TrojAI models are much deeper and have a much larger capacity than our custom CNN models; (iii) TrojAI uses multi-class poisoning compared to the single class poisoning for our CNNs. Yet, given these new challenges, DeBUGCN outperforms the other methods.

    \begin{table*}[h!t]
    \centering
    \resizebox{0.75\textwidth}{!}{
        % \begin{sc}
        \begin{tabular}{c|c|c|c}
            \toprule
            
            \textbf{DeBUGCN Modality} & \textbf{GCN Node feature vector} & \textbf{Train accuracy} & \textbf{Test accuracy} \\
            \midrule
            \midrule
            
            \multicolumn{4}{c}{\textbf{Dataset: MNIST-Static CNNs}} \\
            \midrule
            \multirow{3}{*}{Only FC layer graph} & GCN\_linegraph & 55 & 56 \\
             & GCN\_7 & 96.91 & 98.13 \\
             & GCN\_18 (permuted) & 80 & 94.38 \\
            \midrule
            FC layer graph + Flat graph & GCN\_7 & 96.69 & 98.38 \\
            \midrule
            FC layer graph + 2D graph & GCN\_16a & 96.06 & 98 \\
            \midrule
            \midrule
            
            \multicolumn{4}{c}{\textbf{Dataset: MNIST-Dynamic CNNs}} \\
            \midrule
            \multirow{2}{*}{Only FC layer graph} & GCN\_16a & 99 & 99.75 \\
             & GCN\_18 & 98.63 & 99.75 \\
            \midrule
            FC layer graph + Flat graph & GCN\_18 & 99.06 & 99.88 \\
            \midrule
            FC layer graph + 2D graph & GCN\_16b & 99.22 & 99.90 \\
            
            \bottomrule
        \end{tabular}

        % \end{sc}
    }
    \caption{Table of experiment results of best performing DeBUGCN variant on each MNIST (Static and Dynamic) CNNs dataset.}
    \label{table:best_mnist_experiment_results}
    \end{table*}

    \begin{table*}[h!t]
    \centering
    \resizebox{0.75\textwidth}{!}{
        % \begin{sc}
        \begin{tabular}{c|c|c|c}
            \toprule
            
            \textbf{DeBUGCN Modality} & \textbf{GCN Node feature vector} & \textbf{Train accuracy} & \textbf{Test accuracy} \\
            \midrule
            \midrule
            
            \multicolumn{4}{c}{\textbf{Dataset: CIFAR10-Static CNNs}} \\
            \midrule
            Only FC layer graph & GCN\_16a & 83 & 99.33 \\
            \midrule
            FC layer graph + Flat graph & GCN\_16b & 81.50 & 99.33 \\
            \midrule
            FC layer graph + 2D graph & GCN\_18 & 88.42 & 99.67 \\
            \midrule
            \midrule
            
            \multicolumn{4}{c}{\textbf{Dataset: CIFAR10-Dynamic CNNs}} \\
            \midrule
            Only FC layer graph & GCN\_16b & 84.33 & 99.90 \\
            \midrule
            \multirow{2}{*}{FC layer graph + Flat graph} & GCN\_16a & 83.75 & 98 \\
             & GCN\_18 & 81.58 & 98 \\
            \midrule
            FC layer graph + 2D graph & GCN\_7 & 88.67 & 99.67\\
            
            \bottomrule
        \end{tabular}
        % \end{sc}
    }
    \caption{Table of experiment results of best performing DeBUGCN variant on each CIFAR-10 (Static and Dynamic) CNNs dataset.}
    \label{table:best_cifar10_experiment_results}
    \end{table*}

    \begin{table*}[h!t]
    \centering
    \resizebox{0.75\textwidth}{!}{
        % \begin{sc}
        \begin{tabular}{c|c|c|c}
            \toprule
        
            \textbf{DeBUGCN Modality} & \textbf{GCN Node feature vector} & \textbf{Train accuracy} & \textbf{Test accuracy} \\
            \midrule
            \midrule
            
            \multicolumn{4}{c}{\textbf{Dataset: TrojAI-ResNet CNNs}} \\
            \midrule
            Only FC layer graph & GCN\_16a & 83.53 & 87.59 \\
            \midrule
            FC layer graph + Flat graph & GCN\_16a & 83.35 & 84.83 \\
            \midrule
            FC layer graph + 2D graph & GCN\_18 & 83.18 & 86.67 \\
            \midrule
            \midrule
            
            \multicolumn{4}{c}{\textbf{Dataset: TrojAI-DenseNet CNNs}} \\
            \midrule
            Only FC layer graph & GCN\_18 & 80.47 & 84.06 \\
            \midrule
            FC layer graph + Flat graph & GCN\_16b & 80 & 83.75 \\
            \midrule
            FC layer graph + 2D graph & GCN\_16b & 81.25 & 83.13 \\
            \midrule
            \midrule
            
            \multicolumn{4}{c}{\textbf{Dataset: TrojAI-Full (ResNet + DenseNet) CNNs}} \\
            \midrule
            Only FC layer graph & GCN\_7 & 79.53 & 83.44 \\
            \midrule
            FC layer graph + Flat graph & GCN\_7 & 78.60 & 85.43 \\
            \midrule
            FC layer graph + 2D graph & GCN\_7 & 78.40 & 83.71 \\

            \bottomrule
        \end{tabular}

        % \end{sc}
    }
    \caption{Table of experiment results of best performing DeBUGCN variant on each TrojAI (Static and Dynamic) CNNs dataset.}
    \label{table:best_trojai_experiment_results}
    \end{table*}

\subsection{Experiment 4: Using Convolutional Filters as Graphs}
\label{sec:experiment-conv-graphs}
    
    \begin{figure}[t]
        \centering
        \fbox{\includegraphics[width=1\columnwidth]{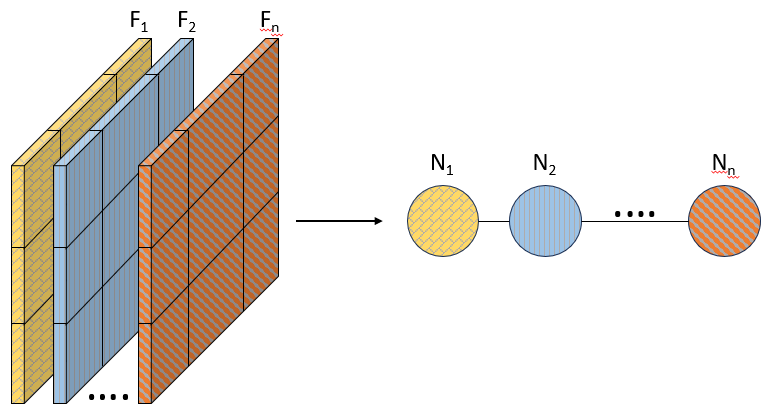}}
        \hfill
        \fbox{\includegraphics[width=1\columnwidth]{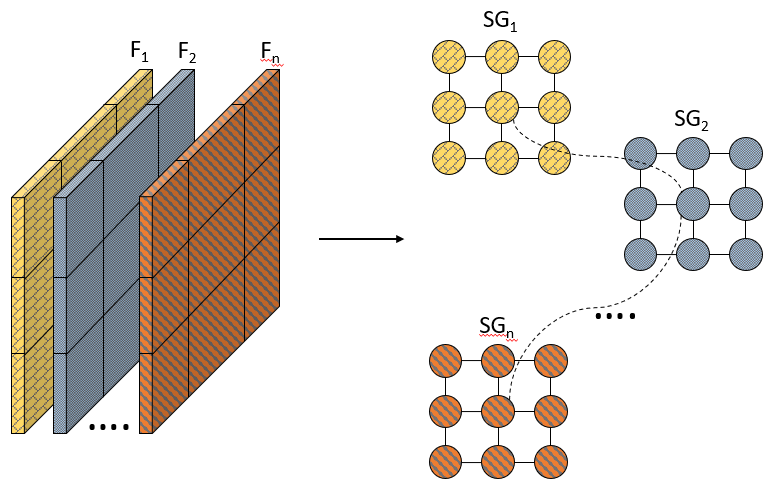}}
        \caption{Examples of convolutional filters as graphs. Top: Flat Graph, Bottom: 2D Graph method.}
        \label{fig:conv_graphs}
    \end{figure}
    
    Following the success of the DeBUGCN pipeline utilizing only the last FC layer weights, we expanded the approach to include convolutional layers.
    This was due to the fact that true real-world application of GCN can only be possible if the GCN can utilize other layers also for classification.
    To this end, we created a multimodal GCN which can use convolutional filters converted to graphs along with FC layer graphs.
    Convolutional layers convey information about image features like colors, shapes and sizes. Early convolutional layers have low level information and this makes them good initial candidates to convert to graphs.
    Convolutional filters have weights and the filters themselves also mimic graph structures. Our hypothesis is that this information can be used by the GCN to improve accuracy.
    We convert the convolutional filters to two types of graphs: flat graphs and 2D graphs.

{\bf Flat Graph method:} 
    We assume each filter to be a single node in the graph and that filters are connected to neighboring filters in series. Each cell in a filter has a learned weight. The corresponding graph node uses a 1-D concatenation of the filter weights as the node's feature vector. 
    Assume a convolutional layer has filter dimensions as $(F_{out}, F_{in}, H, W)$, where $F_{out}$ and $F_{in}$ are the number of output and input channels, and $H$ and $W$ specify the filter sizes.
    The resulting Flat graph has $F_{out}$ nodes and each node has $F_{in}*H*W$ features.
    This is shown in the top image of Fig.~\ref{fig:conv_graphs}.

{\bf 2D Graph method:} 
    For a 2D graph we assume each filter to be a subgraph, which in turn connects to subsequent filters via the center nodes as shown in the bottom image in Fig.~\ref{fig:conv_graphs}. Thus, each filter's 2D shape is retained in the graph. Each cell in a filter has a learned weight. The nodes in the corresponding subgraph now use a feature vector of length 1 which is the cell's learned weight.
    Assume a convolutional layer has filter dimensions as $(F_{out}, F_{in}, H, W)$ as before. The resulting 2D graph has $F_{out}*F_{in}*H*W$ nodes and each node has 1 node feature.

    In this experiment, we used the multimodal version of the GCN.
    as listed in Table~\ref{table:gcn_model_architecture}.
    The first set of 3 GraphConv + mean pool layers operate on the FC layer graphs and the second set performs graph convolutions on the convolutional filter graphs.
    We ran each of the GCNs listed in Table~\ref{table:gcn_notation_feature_vectors} on every dataset. For each run, the input to the GCN is the FC layer graphs (with the corresponding feature vector) + either Flat or 2D filter graphs. The best results using this multimodal GCN pipeline are shown in Table~\ref{table:best_experiment_results}.
    These results show that in almost all datasets, the inclusion of convolution filters as graphs has boosted the GCN's classification accuracy. We saw lower losses and higher accuracies across the board. This gives a strong positive affirmation to our hypothesis that convolutional filters as graphs compensate for useful structural information missing from the FC layer graphs. 
    Looking at the results, it may seem that Flat graphs are better, but in most scenarios the 2D graph was superior because it enables the GCN to maximally utilize the layer's graph topology for message passing.

\subsubsection {Scalability of DeBUGCN}
    Compute intensive trojan detection techniques do not scale because outsourcing raises the possibility of obtaining a trojan model from a third party. As shown in Table \ref{table:computation_time_comparison}, DeBUGCN detection is at least two times faster than the competing trojan detection methods. This is because we only consider the final FC layer's weights forming small graphs with only simple node features. State of the art methods need optimization processes to detect the trojan, hence the high computation time. In AC, network activations are first obtained by feeding example images which again increases the computation time.
    
    \begin{table}[t]
    \centering
    \resizebox{0.65\columnwidth}{!}{
        % \begin{sc}
        \begin{tabular}{c|c|c}
            \toprule
            
            Method & MNIST & TrojAI-Full \\
            
            \midrule
            
            % \hhline{=|=|=}
            DeBUGCN & 321 & {\cellcolor[rgb]{0.929,0.929,0.929}}\textbf{519}  \\
            NC & 1702 & 2919  \\
            ABS  & 1634 & 2820  \\
            ULP & 3292 & 5129  \\
            AC & 612 & 1022 \\
            TABOR & 2945  & 5268  \\
            DLTND & 3189 & 5478 \\

            \bottomrule
        \end{tabular}
        % \end{sc}
    }
    \caption{DeBUGCN's computation time (average, in seconds) compared to other methods.}
    \label{table:computation_time_comparison}
    \end{table}

\subsection{Ablation Study of FC graph's Node Features}
\label{sec:ablation-study}
    We explored the effects of the FC graph's node features on the GCN model's accuracy. The various node feature vectors and their corresponding GCN model's notation are shown in Table~\ref{table:gcn_notation_feature_vectors}.

%\subsubsection{Effect of L/R, node degree}
    The `L/R' feature was useful to the extent that the GCN benefits from knowing whether a node belongs to the left or right group of the bipartite graph.
    This aligns intuitively that a given node in a layer will affect the activations and the weights connecting it to the next layer. 
    This is indicated by the fact that when GCN models did not use the L/R feature, the accuracy was 1-2\% less than the best GCN model.

%\subsubsection{Effect of mean, min, max, and sum}
    The mean, min, max, and sum features are useful because they conveyed aggregated information about a node's connected weights. Weights act as edge features and are important since learned weights reflect which nodes will be activated in the presence of the trigger patch. This is evidenced by a small difference in accuracy of models `GCN\_18' and `GCN\_16a'.

%\subsubsection{Effect of histogram bin counts and histogram bin boundaries}
    We set the number of bins to 5 and calculate the bin counts and bin boundaries of the connected weights of each node.  Histogram bin counts and bin boundary values are clearly the most important feature and revealed the distribution of weight values. 
    In conjunction with the aggregation features, it also showed the effect of any outliers. For example: if a node had a weight distribution in a certain range with an outlier, it could mean that the node was part of the node group that activated in the presence of backdoors. 
    This importance was evidenced by a huge difference in accuracy between models GCN\_16a, GCN\_16b, GCN\_18 vs GCN\_5, GCN\_7. The former set of GCNs consistently achieved higher accuracy compared to the latter on the TrojAI datasets.

%%%%%%%%%%%%%%%%%%%%%%%%%%%%%%%%%%%%%%%%%%%%%%%%%%%%%%%%%%%%%%

\section{Conclusion}
\label{sec:conclusion}
    Our work describes and empirically validates our robust and model-agnostic DeBUGCN pipeline. We hypothesized that CNNs can be presented as graphs and make a case that a GCN model can utilize these graphs to detect backdoored CNNs. 
    We demonstrated the efficacy of DeBUGCN on several datasets and presented an ablation study to examine the importance of various node features. 
    
    Each experiment motivates the next and we carefully examined all results, presented observations and irrefutable evidence to support each claim. 
    MNIST dataset proved feasibility of GCNs in a backdoor detection.
    CIFAR-10 and TrojAI experiments empirically demonstrated DeBUGCN's model-agnostic effectiveness at simultaneously handling multiple architectures.
    Next, Dynamic-* results show that DeBUGCN can detect a trigger's evidence even if dispersed in a graph.
    Lastly, we also compared the accuracy and computation time of our method with other other well-proven and trusted techniques. Our work makes a compelling statement towards using GCNs as trojan detectors. 
    Finally, we also showed that the DeBUGCN pipeline can utilize multiple layer types to make use to network connectivity and topology.
    
    We therefore view this work as a motivation for future research to focus on the utilization of GCNs in detecting other types of attacks on DNN models, including incorporating other layer types and further experimentation of node features. 
    As part of upcoming research, we are exploring how to incorporate the more layer types from other DNNs to create a multi-type multimodal GCN classifier. 
    Past research has shown that knowledge from CNNs can be used by GNNs for classification tasks~\cite{Lu_IEEE2021,Zhang_IEEE2021}.

%%%%%%%%%%%%%%%%%%%%%%%%%%%%%%%%%%%%%%%%%%%%%%%%%%%%%%%%%%%%%%

\section*{Impact Statement}
\label{sec:impact-statement}

This paper aims to advance the field of Adversarial ML with no notable negative societal consequences.

%%%%%%%%%%%%%%%%%%%%%%%%%%%%%%%%%%%%%%%%%%%%%%%%%%%%%%%%%%%%%%

\bibliography{DeBUGCN_paper}
\bibliographystyle{icml2024}

%%%%%%%%%%%%%%%%%%%%%%%%%%%%%%%%%%%%%%%%%%%%%%%%%%%%%%%%%%%%%%%%%%%%%%%%%%%%%%%
%%%%%%%%%%%%%%%%%%%%%%%%%%%%%%%%%%%%%%%%%%%%%%%%%%%%%%%%%%%%%%%%%%%%%%%%%%%%%%%
% APPENDIX
%%%%%%%%%%%%%%%%%%%%%%%%%%%%%%%%%%%%%%%%%%%%%%%%%%%%%%%%%%%%%%%%%%%%%%%%%%%%%%%
%%%%%%%%%%%%%%%%%%%%%%%%%%%%%%%%%%%%%%%%%%%%%%%%%%%%%%%%%%%%%%%%%%%%%%%%%%%%%%%
\newpage
\appendix
\onecolumn
\section{Results Appendix}

    \begin{table*}[h]
    \centering
    \resizebox{0.9\textwidth}{!}{
        % \begin{sc}
        \begin{tabular}{c|c|c|c}
            \toprule
            
            \textbf{GCN Model} & \textbf{Node feature vector} & \textbf{Train accuracy} & \textbf{Test accuracy} \\
            \midrule
            \midrule
            
            \multicolumn{4}{c}{Dataset: MNIST-Static CNNs} \\
            \midrule
             & GCN\_linegraph & 55 & 56 \\
             & GCN\_5 & 95.28 & 94.75 \\
             & GCN\_7 & {\cellcolor[rgb]{0.929,0.929,0.929}}\textbf{96.91} & 98.13 \\
             & GCN\_16a & 95.59 & 96.38 \\
             & GCN\_16b & 94.06 & 96.25 \\
             & GCN\_18 (permuted) & 80.0 & 94.38 \\
            \multirow{-7}{*}{Only FC layer graph} & GCN\_18 & 96.59 & 97.63 \\
            \midrule
             & GCN\_5 & 94.72 & 96.88 \\
             & GCN\_7 & 96.69 & {\cellcolor[rgb]{0.929,0.929,0.929}}\textbf{98.38} \\
             & GCN\_16a & 95.84 & 97.50 \\
             & GCN\_16b & 93.97 & 95.75 \\
            \multirow{-5}{*}{FC layer graph + Flat convolutional filters graph} & GCN\_18 & 96.56 & 96.75 \\
            \midrule
             & GCN\_5 & 95.50 & 96.13 \\
             & GCN\_7 & 95.97 & 97.63 \\
             & GCN\_16a & 96.06 & 98 \\
             & GCN\_16b & 93.72 & 94.63 \\
            \multirow{-5}{*}{FC layer graph + 2D convolutional filters graph} & GCN\_18 & 96.84 & 97 \\
            \midrule
            \midrule
            
            \multicolumn{4}{c}{Dataset: MNIST-Dynamic CNNs} \\
            \midrule
             & GCN\_5 & 98.44 & 99 \\
             & GCN\_7 & 98 & 99.13 \\
             & GCN\_16a & 99 & 99.75 \\
             & GCN\_16b & 99.06 & 99.58 \\
            \multirow{-5}{*}{Only FC layer graph} & GCN\_18 & 98.63 & 99.75 \\
            \midrule
             & GCN\_5 & 98.75 & 98.88 \\
             & GCN\_7 & 98.06 & 98.75 \\
             & GCN\_16a & 99.03 & 99.75 \\
             & GCN\_16b & 98.70 & 99.79 \\
            \multirow{-5}{*}{FC layer graph + Flat convolutional filters graph} & GCN\_18 & 99.06 & 99.88 \\
            \midrule
             & GCN\_5 & 98.61 & 99.03 \\
             & GCN\_7 & 97.25 & 99 \\
             & GCN\_16a & 98.81 & 99.50 \\
             & GCN\_16b & 99.22 & {\cellcolor[rgb]{0.929,0.929,0.929}}\textbf{99.90} \\
            \multirow{-5}{*}{FC layer graph + 2D convolutional filters graph} & GCN\_18 & {\cellcolor[rgb]{0.929,0.929,0.929}}\textbf{99.31} & 99.50 \\
            \midrule
            \midrule
            
            \bottomrule
        \end{tabular}

        % \end{sc}
    }
    \caption{Table of experiment results on MNIST (static and dynamic) CNNs datasets. Highest train and test accuracy of DeBUGCN variant is highlighted.}
    \label{table:mnist_experiment_results}
    \end{table*}

    \begin{table*}[h]
    \centering
    \resizebox{0.9\textwidth}{!}{
        % \begin{sc}
        \begin{tabular}{c|c|c|c}
            \toprule
            
            \textbf{GCN Model} & \textbf{Node feature vector} & \textbf{Train accuracy} & \textbf{Test accuracy} \\
            \midrule
            \midrule
            
            \multicolumn{4}{c}{Dataset: CIFAR10-Static CNNs} \\
            \midrule
             & GCN\_5 & 81.50 & 79 \\
             & GCN\_7 & 88.33 & 98.33 \\
             & GCN\_16a & 83 & 99.33 \\
             & GCN\_16b & 81.25 & 98.33 \\
            \multirow{-5}{*}{Only FC layer graph} & GCN\_18 & 84.92 & 97.33 \\
            \midrule
             & GCN\_5 & 79.50 & 82.67 \\
             & GCN\_7 & 87.75 & 94 \\
             & GCN\_16a & 88.08 & 99 \\
             & GCN\_16b & 81.50 & 99.33 \\
            \multirow{-5}{*}{FC layer graph + Flat convolutional filters graph} & GCN\_18 & 84.42 & 98.33 \\
            \midrule
             & GCN\_5 & 79.92 & 87.33 \\
             & GCN\_7 & {\cellcolor[rgb]{0.929,0.929,0.929}}\textbf{88.50} & 98.67 \\
             & GCN\_16a & 86.75 & 99 \\
             & GCN\_16b & 85.25 & 99 \\
            \multirow{-5}{*}{FC layer graph + 2D convolutional filters graph} & GCN\_18 & 88.42 & {\cellcolor[rgb]{0.929,0.929,0.929}}\textbf{99.67} \\
            \midrule
            \midrule
            
            \multicolumn{4}{c}{Dataset: CIFAR10-Dynamic CNNs} \\
            \midrule
             & GCN\_5 & 82.33 & 82.33 \\
             & GCN\_7 & 87.92 & 95 \\
             & GCN\_16a & 85.75 & 97.67 \\
             & GCN\_16b & 84.33 & {\cellcolor[rgb]{0.929,0.929,0.929}}\textbf{99.90} \\
            \multirow{-5}{*}{Only FC layer graph} & GCN\_18 & 83.50 & 96.67 \\
            \midrule
             & GCN\_5 & 82.58 & 81 \\
             & GCN\_7 & 87.00 & 93.67 \\
             & GCN\_16a & 83.75 & 98.00 \\
             & GCN\_16b & 84.33 & 97.33 \\
            \multirow{-5}{*}{FC layer graph + Flat convolutional filters graph} & GCN\_18 & 81.58 & 98 \\
            \midrule
             & GCN\_5 & 82.42 & 81.67 \\
             & GCN\_7 & {\cellcolor[rgb]{0.929,0.929,0.929}}\textbf{88.67} & 99.67 \\
             & GCN\_16a & 87.83 & 99.33 \\
             & GCN\_16b & 86.08 & 99.33 \\
            \multirow{-5}{*}{FC layer graph + 2D convolutional filters graph} & GCN\_18 & 84.67 & 98 \\
            
            \bottomrule
        \end{tabular}
        % \end{sc}
    }
    \caption{Table of experiment results on CIFAR-10 (static and dynamic) CNNs datasets. Highest train and test accuracy of DeBUGCN variant is highlighted.}
    \label{table:cifar10_experiment_results}
    \end{table*}

    \begin{table*}[h]
    \centering
    \resizebox{0.85\textwidth}{!}{
        % \begin{sc}
        \begin{tabular}{c|c|c|c}
            \toprule
        
            \textbf{GCN Model} & \textbf{Node feature vector} & \textbf{Train accuracy} & \textbf{Test accuracy} \\
            \midrule
            \midrule
            
            \multicolumn{4}{c}{Dataset: TrojAI-ResNet CNNs} \\
             & GCN\_5 & 63.64 & 59.54 \\
             & GCN\_7 & 68.09 & 71.95 \\
             & GCN\_16a & 83.53 & {\cellcolor[rgb]{0.929,0.929,0.929}}\textbf{87.59} \\
             & GCN\_16b & 84.34 & 84.83 \\
            \multirow{-5}{*}{Only FC layer graph} & GCN\_18 & 83.58 & 85.29 \\
            \midrule
             & GCN\_5 & 62.60 & 63.68 \\
             & GCN\_7 & 67.86 & 72.18 \\
             & GCN\_16a & 83.35 & 84.83 \\
             & GCN\_16b & {\cellcolor[rgb]{0.929,0.929,0.929}}\textbf{84.91} & 81.84 \\
            \multirow{-5}{*}{FC layer graph + Flat convolutional filters graph} & GCN\_18 & 84.57 & 84.60 \\
            \midrule
             & GCN\_5 & 62.66 & 63.68 \\
             & GCN\_7 & 70.75 & 71.26 \\
             & GCN\_16a & 83.99 & 84.60 \\
             & GCN\_16b & 83.70 & 85.75 \\
            \multirow{-5}{*}{FC layer graph + 2D convolutional filters graph} & GCN\_18 & 83.18 & 86.67 \\
            \midrule
            \midrule
            
            \multicolumn{4}{c}{Dataset: TrojAI-DenseNet CNNs} \\
            \midrule
             & GCN\_5 & 60.16 & 58.44 \\
             & GCN\_7 & 73.88 & 70 \\
             & GCN\_16a & 80.39 & 83.44 \\
             & GCN\_16b & {\cellcolor[rgb]{0.929,0.929,0.929}}\textbf{81.57} & 83.13 \\
            \multirow{-5}{*}{Only FC layer graph} & GCN\_18 & 80.47 & {\cellcolor[rgb]{0.929,0.929,0.929}}\textbf{84.06} \\
            \midrule
             & GCN\_5 & 61.41 & 60.63 \\
             & GCN\_7 & 71.76 & 77.19 \\
             & GCN\_16a & 80.39 & 82.50 \\
             & GCN\_16b & 80 & 83.75 \\
            \multirow{-5}{*}{FC layer graph + Flat convolutional filters graph} & GCN\_18 & 80.31 & 81.88 \\
            \midrule
             & GCN\_5 & 59.37 & 59.38 \\
             & GCN\_7 & 73.41 & 75.31 \\
             & GCN\_16a & 81.49 & 82.50 \\
             & GCN\_16b & 81.25 & 83.13 \\
            \multirow{-5}{*}{FC layer graph + 2D convolutional filters graph} & GCN\_18 & 80.24 & 81.25 \\
            \midrule
            \midrule
            
            \multicolumn{4}{c}{Dataset: TrojAI-Full (ResNet + DenseNet) CNNs} \\
            \midrule
             & GCN\_5 & 72.75 & 72.05 \\
             & GCN\_7 & {\cellcolor[rgb]{0.929,0.929,0.929}}\textbf{79.53} & 83.44 \\
             & GCN\_16a & 78.34 & 80.13 \\
             & GCN\_16b & 78.24 & 82.25 \\
            \multirow{-5}{*}{Only FC layer graph} & GCN\_18 & 78.30 & 80.79 \\
            \midrule
             & GCN\_5 & 74.58 & 73.38 \\
             & GCN\_7 & 78.60 & {\cellcolor[rgb]{0.929,0.929,0.929}}\textbf{85.43} \\
             & GCN\_16a & 75.77 & 80.66 \\
             & GCN\_16b & 76.94 & 82.12 \\
            \multirow{-5}{*}{FC layer graph + Flat convolutional filters graph} & GCN\_18 & 76.54 & 84.64 \\
            \midrule
             & GCN\_5 & 74.38 & 70.60 \\
             & GCN\_7 & 78.40 & 83.71 \\
             & GCN\_16a & 79.40 & 82.78 \\
             & GCN\_16b & 77.01 & 83.31 \\
            \multirow{-5}{*}{FC layer graph + 2D convolutional filters graph} & GCN\_18 & 78.20 & 83.31 \\

            \bottomrule
        \end{tabular}

        % \end{sc}
    }
    \caption{Table of experiment results on TrojAI datasets. Highest train and test accuracy of DeBUGCN variant on each dataset is highlighted.}
    \label{table:trojai_experiment_results}
    \end{table*}
    
%%%%%%%%%%%%%%%%%%%%%%%%%%%%%%%%%%%%%%%%%%%%%%%%%%%%%%%%%%%%%%%%%%%%%%%%%%%%%%%
%%%%%%%%%%%%%%%%%%%%%%%%%%%%%%%%%%%%%%%%%%%%%%%%%%%%%%%%%%%%%%%%%%%%%%%%%%%%%%%

\end{document}